\documentclass[conference]{IEEEtran}

\usepackage{copyright}

\usepackage[english]{babel}

\usepackage{amssymb}
\usepackage{amsmath}
\usepackage{hyperref}
\usepackage{cleveref}
\usepackage{graphicx}
\usepackage{latexsym}
\usepackage{algorithm}
\usepackage{algpseudocode}
\usepackage{todonotes}
\usepackage{newclude}
\usepackage{multicol}
\usepackage{tabularx}
\usepackage{booktabs}
\usepackage{threeparttable}

\usepackage{enumitem}
\usepackage{listings}
\usepackage{combelow}
\usepackage{cite}
\usepackage{bigfoot}

\usepackage[labelformat=simple]{subcaption}

\usepackage[acronym]{glossaries}
\newacronym{fmcw}{FMCW}{Frequency-Modulated Continuous-Wave}
\newacronym{fov}{FoV}{field of view}
\newacronym{pr}{PR}{Precision-Recall}
\newacronym{vo}{VO}{Visual Odometry}
\newacronym{gps}{GPS}{Global Positioning System}
\newacronym{auc}{AUC}{area-under-curve}
\newacronym{cnn}{CNN}{Convolutional Neural Network}
\newacronym{dl}{DL}{Deep Learning}
\newacronym{fcnn}{FCNN}{Fully Convolutional Neural Network}
\newacronym{lidar}{LiDAR}{Light Detection and Ranging}
\newacronym{nn}{NN}{nearest neighbour}
\newacronym{lcd}{LCD}{Loop Closure Detection}
\newacronym{roc}{ROC}{Receiver Operating Characteristic}
\newacronym{tr}{TR}{teach-and-repeat}
\newacronym{gan}{GAN}{Generative Adversarial Network}

\usepackage[binary-units]{siunitx}
\usepackage{xcolor}

\usepackage{url}
\usepackage{hyperref}
\newcommand\rurl[1]{%
  \href{http://#1}{\nolinkurl{#1}}%
}

\usepackage{cleveref}
\crefname{table}{Table}{Tables}
\crefname{figure}{Figure}{Figures}
\crefname{section}{Section}{Sections}

\newcommand{\vanilla}{{\sc vgg-16/netvlad}}
\newcommand{\sslam}{{\sc SeqSLAM}}
\newcommand{\nn}{{\sc NN}}
\newcommand{\kradar}{{\sc kRadar}}
\newcommand{\rslam}{{\sc LAY}}

\begin{document}

\title{Look Around You: Sequence-based Radar Place Recognition with Learned Rotational Invariance}
\author{Matthew Gadd, Daniele De Martini, and Paul Newman\\
Oxford Robotics Institute, Dept. Engineering Science, University of Oxford, UK.\\\texttt{\{mattgadd,daniele,pnewman\}@robots.ox.ac.uk}}
\maketitle

\copyrightnotice

\begin{abstract}
This paper details an application which yields significant improvements to the adeptness of place recognition with \acrlong{fmcw} radar -- a commercially promising sensor poised for exploitation in mobile autonomy.
We show how a rotationally-invariant metric embedding for radar scans can be integrated into sequence-based trajectory matching systems typically applied to videos taken by visual sensors.
Due to the complete horizontal \acrlong{fov} inherent to the radar scan formation process, we show how this off-the-shelf sequence-based trajectory matching system can be manipulated to detect place matches when the vehicle is travelling down a previously visited stretch of road in the opposite direction.
We demonstrate the efficacy of the approach on \SI{26}{\kilo\metre} of challenging urban driving taken from the largest radar-focused urban autonomy dataset released to date -- showing a boost of \SI{30}{\percent} in recall at high levels of precision over a \acrlong{nn} approach.
\end{abstract}


\section{Introduction}

In order for autonomous vehicles to travel safely at higher speeds or operate in wide-open spaces where there is a dearth of distinct features, a new level of robust sensing is required.
\gls{fmcw} radar satisfies these requirements, thriving in all environmental conditions (rain, snow, dust, fog, or direct sunlight), providing a \SI{360}{\degree} view of the scene, and detecting targets at ranges of up to hundreds of metres with centimetre-scale precision.
Indeed, there is a burgeoning interest in exploiting \gls{fmcw} radar to enable robust mobile autonomy, including ego-motion estimation~\cite{cen2018precise,cen2019radar,2019ICRA_aldera,2019ITSC_aldera,Barnes2019MaskingByMoving,UnderTheRadarArXiv}, localisation~\cite{KidnappedRadarArXiv,tang2020rsl}, and scene understanding~\cite{weston2019probably}.

\begin{figure}
    \centering
    \includegraphics[width=\columnwidth]{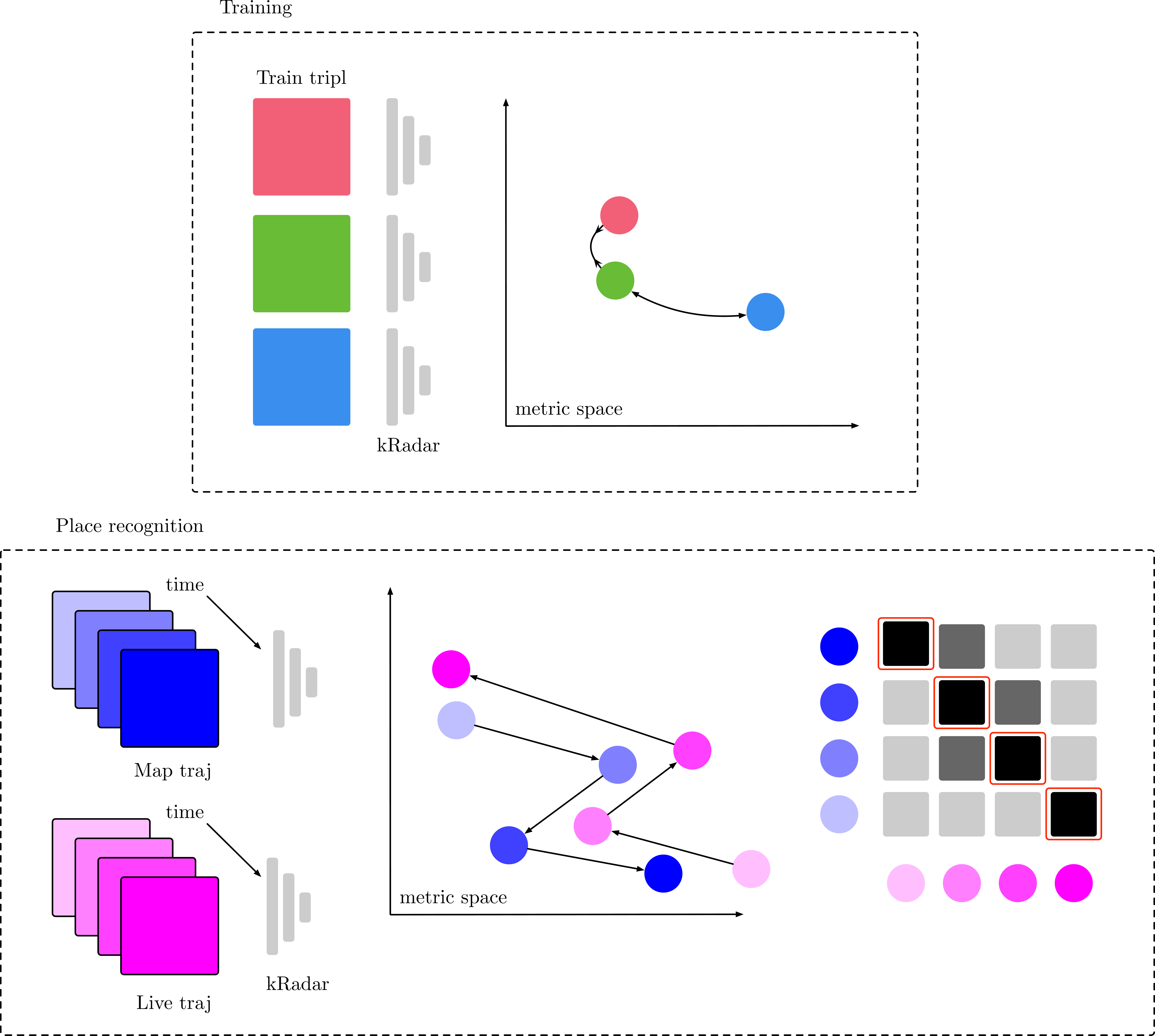}
    \caption{An overview of our pipeline.
    The offline stages include \emph{enforcing} a metric space by training a \gls{fcnn} which takes polar radar scans as input, and \emph{encoding} a trajectory of scans (the map) by forward passes through this network (c.f.~\cref{sec:method:kradar}).
    The online stages involve \emph{inference} to represent the place the robot currently finds itself within in terms of the learned knowledge and \emph{querying} the space (c.f.~\cref{sec:method:seqslam}) which -- in contrast to our prior work -- involves a search for coherent sequences of matches rather than a globally closest frame in the embedding space.}
    \label{fig:pipeline}
    \vspace{-.5cm}
\end{figure}

\cref{fig:pipeline} shows an overview of the pipeline proposed in this paper which extends our recent work in extremely robust radar-only place recognition~\cite{KidnappedRadarArXiv} in which a metric space for embedding polar radar scans was learned, facilitating topological localisation using \gls{nn} matching.
We show that this learned metric space can be leveraged within a sequenced-based topological localisation framework to bolster matching performance by both mitigating visual similarities that are caused by the planarity of the sensor and failures due to sudden obstruction in dynamic environments.
Due to the complete horizontal \gls{fov} of the radar scan formation process, we show how the off-the-shelf sequence-based trajectory matching system can be manipulated to allow us to detect place matches when the vehicle is travelling down a previously visited stretch of road in the opposite direction.

This paper proceeds by reviewing related literature in~\cref{sec:rel_work}.
\cref{sec:method} describes our approach for a more canny use of a metric space in which polar radar scans are embedded.
We describe in~\cref{sec:experimental} details for implementation, evaluation, and our dataset.
\cref{sec:results} discusses results from such an evaluation.
\cref{sec:concl,sec:fut} summarise the findings and suggest further avenues for investigation.

\section{Related Work}
\label{sec:rel_work}

Recent work has shown the promise of \gls{fmcw} radar for robust place recognition~\cite{KidnappedRadarArXiv,gskim2020mulran} and metric localisation~\cite{UnderTheRadarArXiv}.
None of these methods account for temporal effects in the radar measurement stream.

SeqSLAM~\cite{milford2012seqslam} and its variants have been extremely successful at tackling large-scale, robust place recognition with video imagery in the last decade.
Progress along these lines has included automatic scaling for viewpoint invariance~\cite{pepperell2015automatic}, probabilistic adjustments to the search technique~\cite{hansen2014visual}, and dealing with challenging visual appearance change using \glspl{gan}~\cite{latif2018addressing}.

The work presented in this paper is most closely influenced by the use of feature embeddings learned by training \glspl{cnn}~\cite{dongdong2018cnn}, omnidirectional cameras~\cite{cheng2019panoramic}, and \gls{lidar}~\cite{yin2018synchronous} within the SeqSLAM framework.

\section{Methodology}
\label{sec:method}

Broadly, our method can be summarised as leveraging very recent results in \gls{dl} techniques which provide good metric embeddings for the global location of radar scans within a robust sequence-based trajectory matching system.
We begin the discussion with a brief overview of the baseline SeqSLAM algorithm, followed by a light description of the learned metric space, and concluded by an application which unifies these systems -- the main contribution of this paper.

\subsection{Overview of SeqSLAM}
\label{sec:method:seqslam}

Our implementation of the proposed system is based on an open-source, publicly available port of the original algorithm\footnote{\rurl{https://github.com/tmadl/pySeqSLAM}}.

Incoming images are preprocessed by downsampling (to thumbnail resolution) and patch normalisation.
A difference matrix is constructed storing the euclidean distance between all image pairs.
This difference matrix is then contrast enhanced.
Examples of these matrices can be seen in~\cref{fig:diff_m}.
For more detail, a good summary is available in~\cite{sunderhauf2013we}.

When looking for a match to a query image, SeqSLAM sweeps through the contrast-enhanced difference matrix to find the best matching sequence of adjacent frames.

In the experiments (c.f.~\cref{sec:experimental,sec:results}) we refer to this baseline search as~\sslam.

\begin{figure}
    \centering
    \includegraphics[width=0.8\columnwidth]{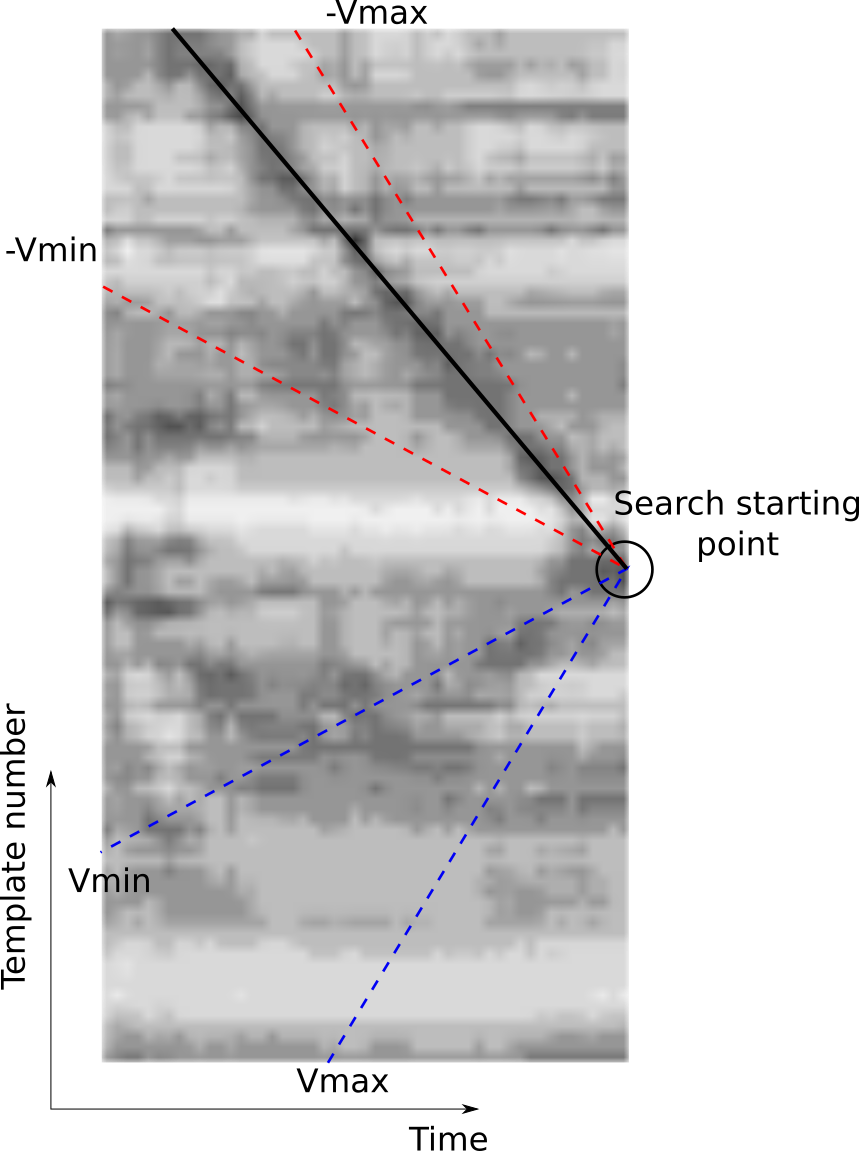}
    \caption{The off-the-shelf sequence matching SeqSLAM system is manipulated in this paper to facilitate backwards \gls{lcd}.
    This is achieved by mirroring the set, $v_{min} < v < v_{max}$ (blue), of trajectories considered -- also considering $-v_{max} < v < -v_{min}$ (red).
    Importantly, this is not a useful adjustment under a na\"{i}ve application of SeqSLAM to radar images and is only beneficial if a rotationally invariant representation is used to construct the difference matrices.}
    \label{fig:seqslam_bw_search}
\end{figure}

\subsection{Overview of Kidnapped Radar}
\label{sec:method:kradar}

To learn filters and cluster centres which help distinguish polar radar images for place recognition we use 
NetVLAD~\cite{arandjelovic2016netvlad} with VGG-16~\cite{simonyan2014very} as a front-end feature extractor -- both popularly applied to the place recognition problem.
Importantly, we make alterations such that the network invariant to the orientation of input radar scans, including: circular padding~\cite{wang2018omnidirectional}, anti-aliasing blurring~\cite{zhang2019making} azimuth-wise max-pooling.

To enforce the metric space, we perform online triplet mining and apply the triplet loss described in~\cite{schroff2015facenet}.
Loop closure labels are taken from a ground truth dataset (c.f.~\cref{sec:experimental}).

The interested reader is referred to~\cite{KidnappedRadarArXiv} for more detail.

In the experiments (c.f.~\cref{sec:experimental,sec:results}) we refer to representations obtained in this manner as~\kradar.

\subsection{Sequence-based Radar Place Recognition}
\label{sec:method:rseqslam}

We replace the image preprocessing step with inference on the network described in~\cref{sec:method:kradar}, resulting in radar scan descriptors of size \num{4096}.

The difference matrix is obtained by calculating the Euclidean distance between every pair of embeddings taken from places along the reference and live trajectories in a window of length $W$.
This distance matrix is then locally contrast enhanced in sections of length $R$.

When searching for a match to a query image, we perform a sweep through this contrast-enhanced difference matrix to find the best matching sequence of frames based on the sum of sequence differences.
In order to be able to detect matches in reverse, this procedure is repeated with a time-reversed live trajectory -- this method would not be applicable to narrow \gls{fov} cameras but is appropriate here as the radar has a \SI{360}{\degree} \gls{fov}.

A simple visualisation of the process is shown in~\cref{fig:seqslam_bw_search}.
In tis case, the forward search (blue lines) is mirrored to perform a backwards search (red lines), which results in the selection of the best match (solid black line).
In the experiments (c.f.~\cref{sec:experimental,sec:results}) we refer to this modified search as~\rslam~(``Look Around You'').

\begin{figure*}
\centering
\begin{subfigure}{0.24\textwidth}
\includegraphics[width=\textwidth]{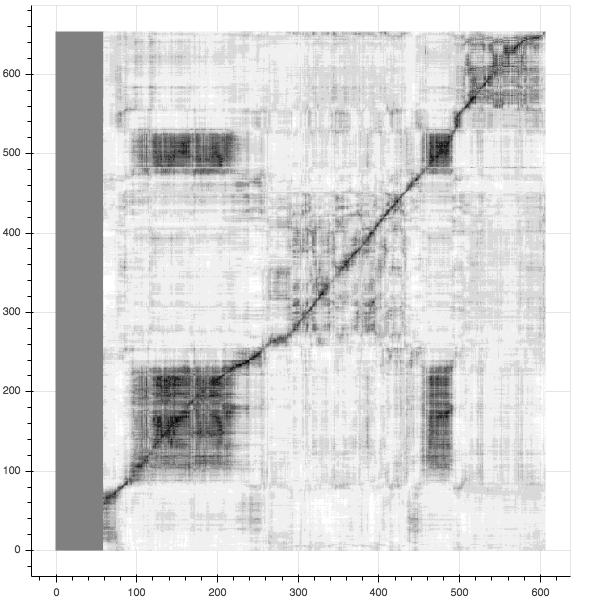}
\caption{}
\label{fig:diff_m_noenh_norot}
\end{subfigure}
\begin{subfigure}{0.24\textwidth}
\includegraphics[width=\textwidth]{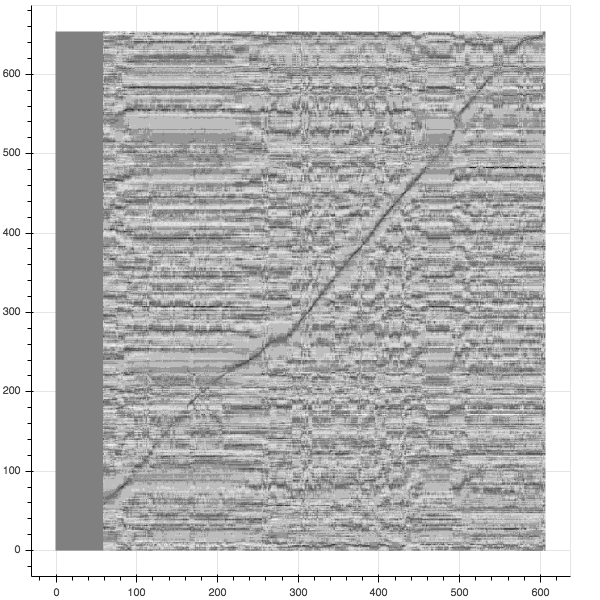}
\caption{}
\label{fig:diff_m_enh_norot}
\end{subfigure}
\begin{subfigure}{0.24\textwidth}
\includegraphics[width=\textwidth]{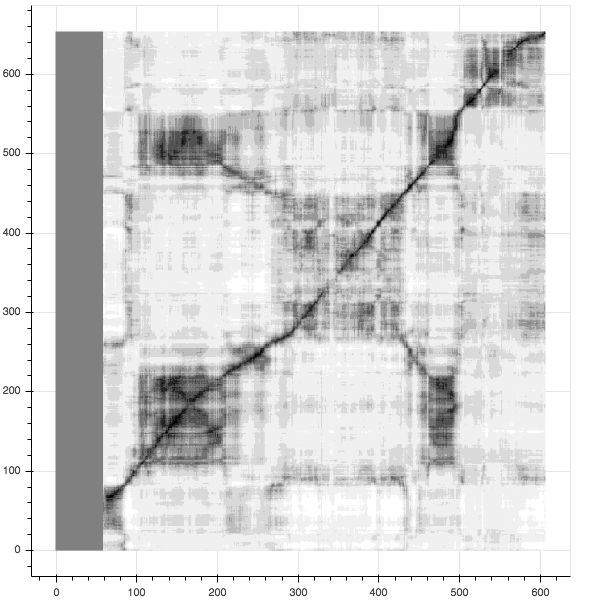}
\caption{}
\label{fig:diff_m_noenh}
\end{subfigure}
\begin{subfigure}{0.24\textwidth}
\includegraphics[width=\textwidth]{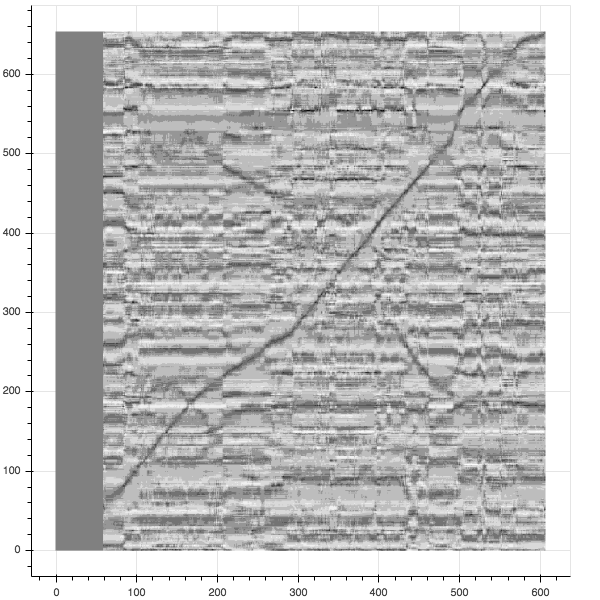}
\caption{}
\label{fig:diff_m_enh}
\end{subfigure}

\begin{subfigure}{0.24\textwidth}
\includegraphics[width=\textwidth]{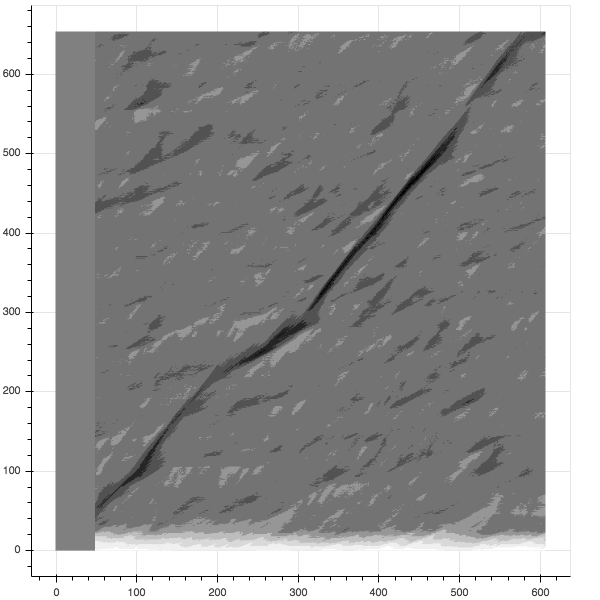}
\caption{}
\label{fig:scores_fwd_vanilla}
\end{subfigure}
\begin{subfigure}{0.24\textwidth}
\includegraphics[width=\textwidth]{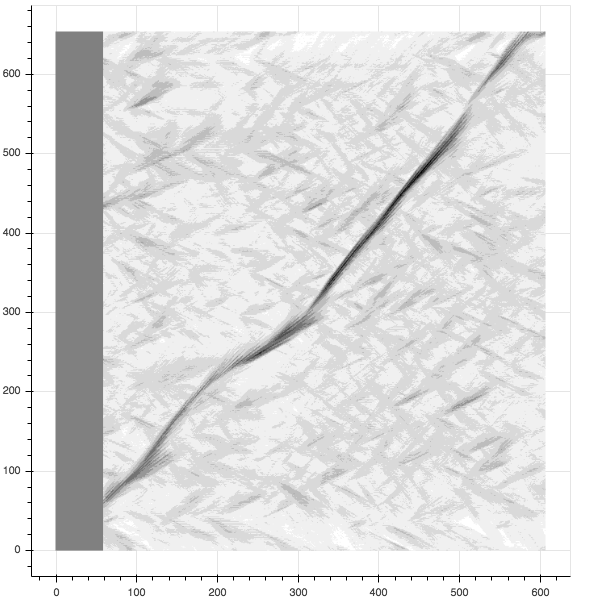}
\caption{}
\label{fig:scores_bwd_vanilla}
\end{subfigure}
\begin{subfigure}{0.24\textwidth}
\includegraphics[width=\textwidth]{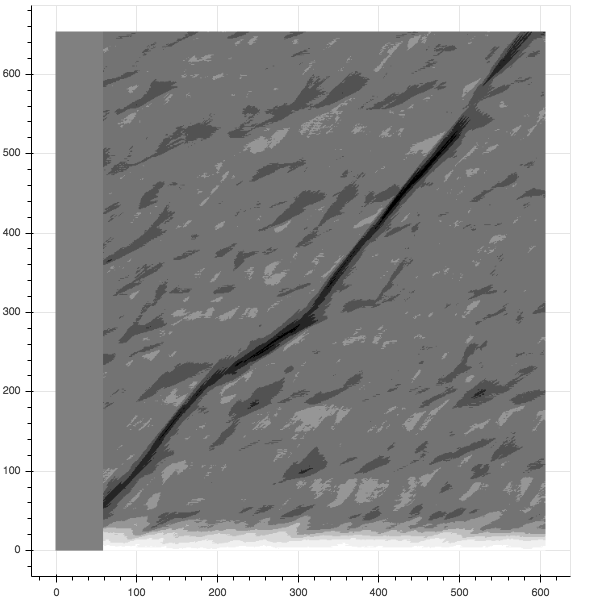}
\caption{}
\label{fig:scores_fwd_ours}
\end{subfigure}
\begin{subfigure}{0.24\textwidth}
\includegraphics[width=\textwidth]{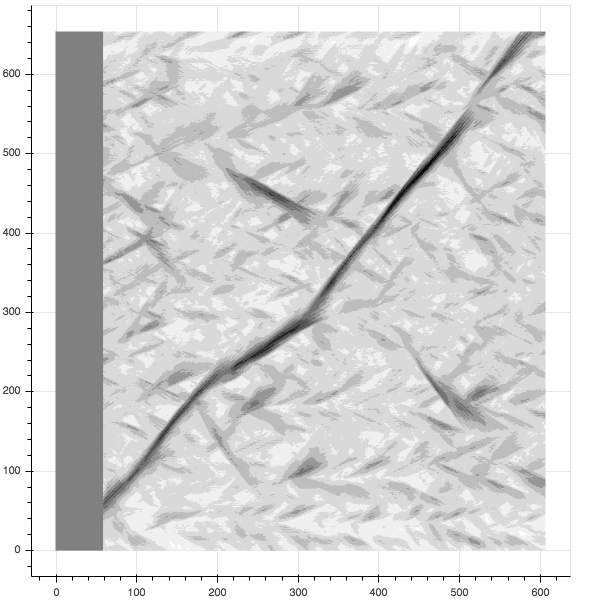}
\caption{}
\label{fig:scores_bwd_ours}
\end{subfigure}
\caption{Difference matrices upon which the SeqSLAM variants perform trajectory searches (top row) and relative match-score-matrices (bottom row).
These are constructed matching the representation of radar scans in two trajectories (rows-versus-columns for each matrix) -- \vanilla~on the left side (a, b, e and f) and~\kradar~on the right side (c, d, g and h).
(a) and (c) are the difference matrices before enhancement -- directly used by the \gls{nn} search in~\cite{KidnappedRadarArXiv} -- and (b) and (d) are the respective enhanced form -- on which SeqSLAM performs its searches.
(e) and (f) are constructed using embeddings inferred by \vanilla~in the enhanced form (b), while (g) and (h) use embeddings inferred by \kradar~in the enhanced form (d).
(e) and (g) are computed by using the forward-style match score method employed by standard SeqSLAM; in contrast, (f) and (h) employ the proposed backward-style match score method.
All match-score matrices are not defined for the first window of columns as SeqSLAM must fill a buffer of frames before any matching is possible.}
\label{fig:diff_and_scores}
\end{figure*}

This procedure is performed for each template, on which a threshold is applied to select the best matches.
\Cref{sec:results} discusses the application of the threshold and reports the results in comparison to the original SeqSLAM approach; in particular \cref{fig:diff_and_scores} shows visual examples of the discussed methodology.

\section{Experimental Setup}
\label{sec:experimental}

This section details our experimental design in obtaining the results to follow in~\cref{sec:results}.

\subsection{Vehicle and radar specifications}

Data was collected using the \textit{Oxford RobotCar} platform~\cite{RobotCarDatasetIJRR}.
The vehicle, as described in the \textit{Oxford Radar RobotCar Dataset}~\cite{RadarRobotCarDatasetArXiv}, is fitted with a CTS350-X Navtech \gls{fmcw} scanning radar.

\subsection{Ground truth database}

The ground truth database is curated offline to capture the sets of nodes that are at a maximum distance (\SI{15}{\metre}) from a query frame, creating a graph-structured database that yields triplets of nodes for training the representation discussed in~\cref{sec:method:kradar}.

To this end, we adjust the accompanying ground truth odometry described in~\cite{RadarRobotCarDatasetArXiv} in order to build a database of ground truth locations.
We manually selected a moment during which the vehicle was stationary at a common point and trimmed each ground trace accordingly.
We also aligned the ground traces by introducing a small rotational offset to account for differing attitudes.

\begin{figure}
    \centering
    \includegraphics[width=0.7\columnwidth]{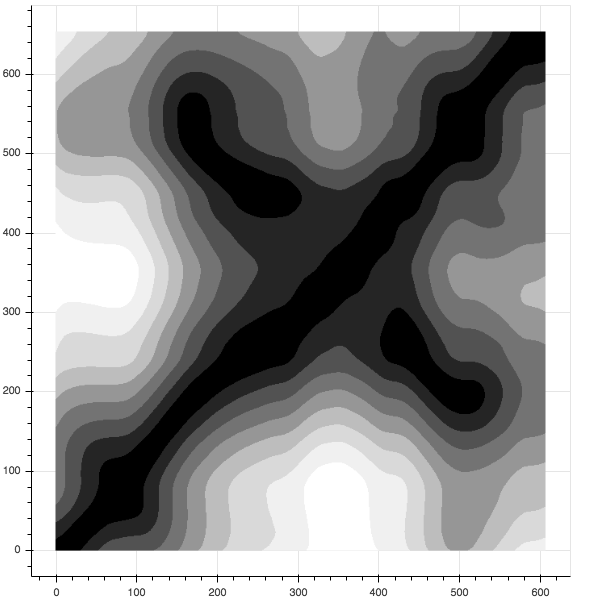}
    \caption{Visualisation of a ground truth $SE(2)$ matrix showing the Euclidean distance between the global positiosn that pairs of radar scans were captured at.
    Each trajectory pair is associated with such a matrix.
    Values in these matrices are scaled from distant (white) to nearby (black).
    In this region of the dataset, the vehicle revisits the same stretch of the route in the opposite direction -- visible as the contours perpendicular to the main diagonal.
    \label{fig:gt}}
\end{figure}

\subsection{Trajectory reservation}
\label{sec:experimental:demarc}

Each approximately \SI{9}{\kilo\metre} trajectory in the Oxford city centre was divided into three distinct portions: \textit{train}, \textit{valid}, and \textit{test}.

The network is trained with ground truth topological localisations between two reserved trajectories in the \emph{train} split.

The \textit{test} split, upon which the results presented in~\cref{sec:results} are based, was specifically selected to feature vehicle traversals over portions of the route in the opposite direction; data from this split are not seen by the network during training.

The results focus on a \gls{tr} scenario, in which all remaining trajectories in the dataset are localised against a map built from the first trajectory that we did not use for learning, totalling \num{27} trajectory pairs (and \SI{26}{\kilo\metre} of driving) with the same map but a different localisation run.

\subsection{Measuring performance}
\label{sec:experimental:metrics}

In the ground truth $SE(2)$ database, all locations within a \SI{15}{\metre} radius of a ground truth location are considered true positives whereas those outside are considered true negatives, a more strictly imposed boundary than in~\cite{KidnappedRadarArXiv}.

Evaluation of \acrfull{pr} is different for the sequence- and \gls{nn}-based approaches.
For the \gls{nn}-based approach of~\cite{KidnappedRadarArXiv} we perform a ball search of the discretised metric space out to a varying embedding distance threshold.
For the sequence-based approach advocated in this paper, we vary the minimum match score.

As useful summaries of \gls{pr} performance, we analyse \gls{auc} as well as some F-scores, including $F_{1}$, $F_{2}$, and $F_{\beta}$ with $\beta = 0.5$~\cite{pino1999modern}.

\begin{figure*}
\centering
\begin{subfigure}{0.35\textwidth}
\includegraphics[width=\textwidth]{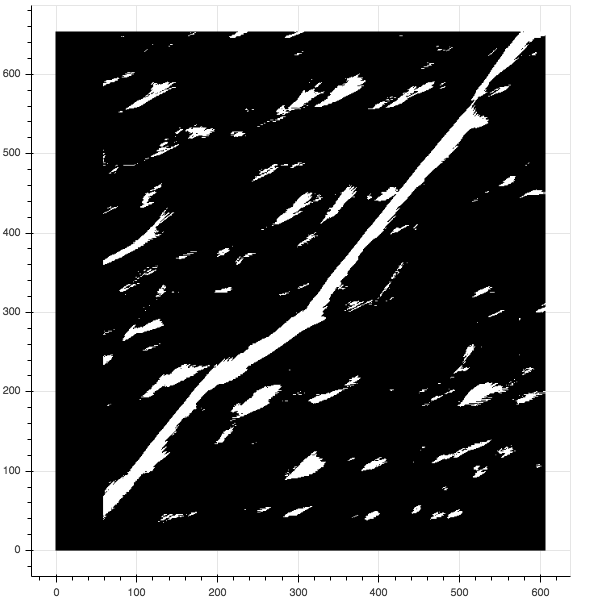}
\caption{}
\label{fig:score_fwd_thr}
\end{subfigure}
\begin{subfigure}{0.35\textwidth}
\includegraphics[width=\textwidth]{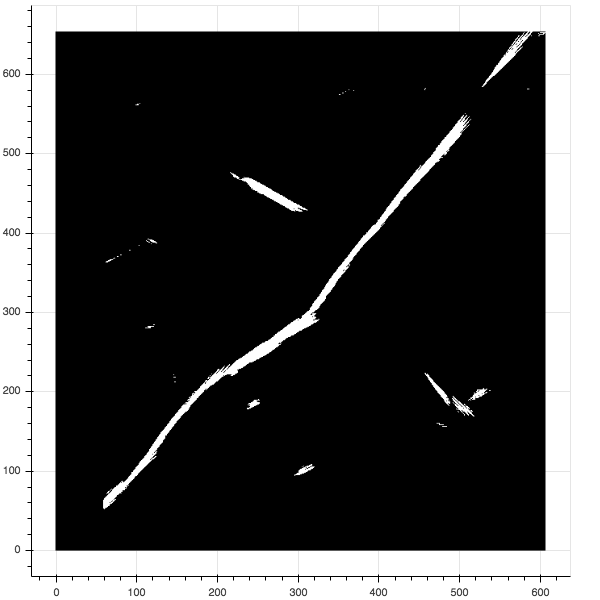}
\caption{}
\label{fig:score_bwd_thr}
\end{subfigure}
\caption{Binarised match score matrices for the \subref{fig:score_fwd_thr} baseline and \subref{fig:score_bwd_thr} mirrored SeqSLAM variants.
The threshold applied for binarisation is higher for~\subref{fig:score_fwd_thr} (\kradar,\sslam) than for (\kradar,\rslam).
This is in order to qualitatively show that even when increasing numbers of potential matches are allowed in \sslam~(high recall), the true backwards loop closures are not featured.
For \rslam~(right), they are.
From these it is evident that the tailored changes to the fundamental SeqSLAM search strategy are better suited to discover loop closures as the vehicle revisits the same route section with opposing orientation -- a common scenario in structured, urban driving.
\label{fig:score_thr}}
\end{figure*}

\begin{figure}
    \centering
    \includegraphics[width=\columnwidth]{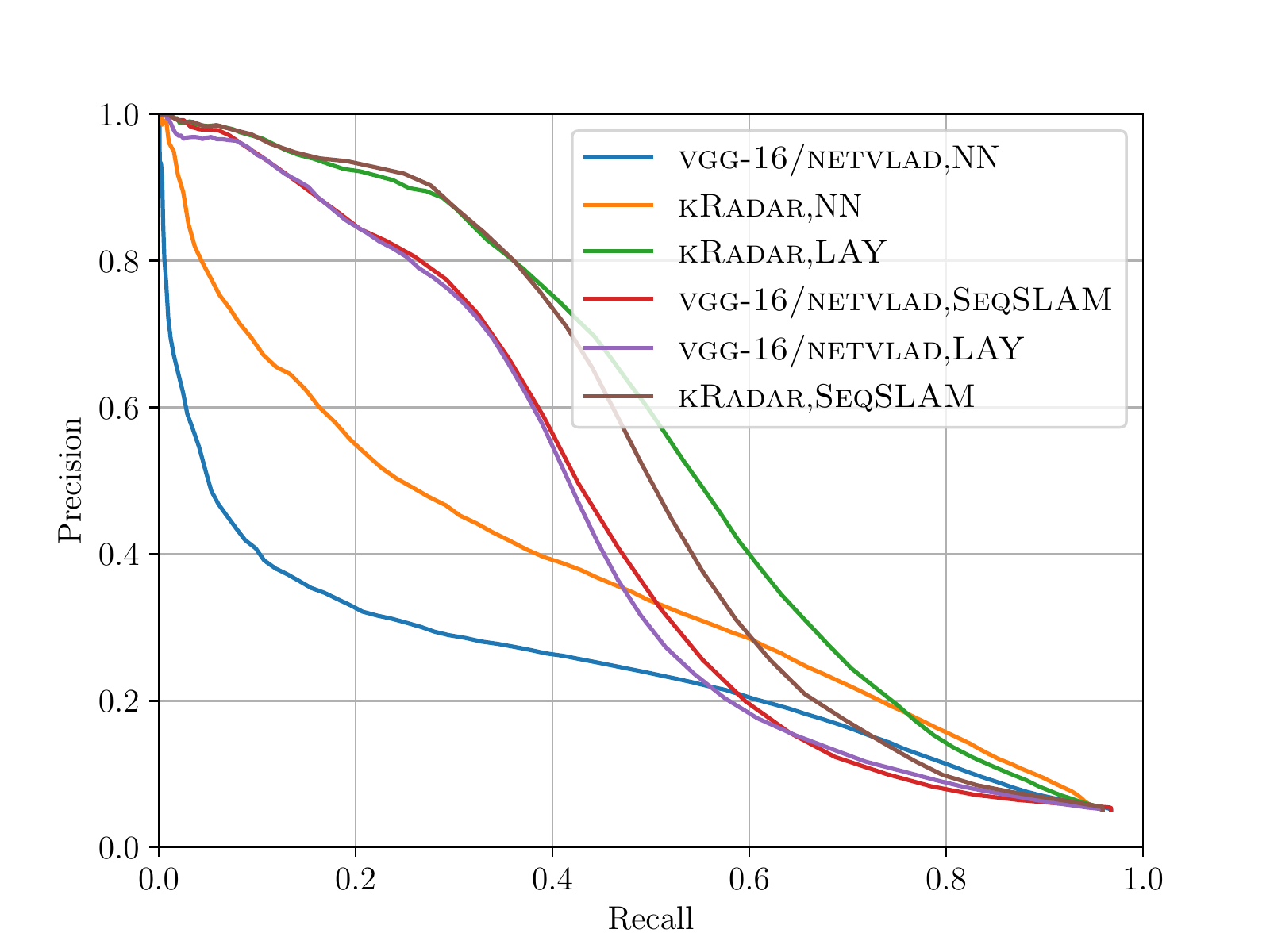}
    \caption{\gls{pr} curves showing the benefit of, firstly, using learned metric embeddings as opposed to radar scans directly and, secondly, the tailored changes to the baseline SeqSLAM search algorithm when performing sequence-based radar place recognition.}
    \label{fig:pr_curves}
\end{figure}

\subsection{Hyperparameter tuning}
\label{sec:experimental:tune}

To produce a fair comparison of the different configurations we can utilise to solve the topological localisation problem, we performed a hyperparameter tuning on the various algorithms; we selected two random trials and excluded them fro the final evaluation.
The window width for the trajectory evaluation $W$ and the enhancement window $R$ have been chosen through a grid search procedure.
The final values are the ones which produced precision-recall curves with the highest value of precision at \SI{80}{\%} recall.

\section{Results}
\label{sec:results}

This section presents instrumentation of the metrics discussed in~\cref{sec:experimental:metrics}.

The hyperparameter optimisation (c.f.~\cref{sec:experimental:tune}) results in the parametisation of the systems for comparison as enumerated in~\cref{tab:hyperpar}.

\begin{table}[]
\renewcommand{\arraystretch}{2}
    \centering
    \begin{tabular}{cc|cc}
         \textbf{Representation} & \textbf{Search} & $R$ & $W$ \\
         \hline
         \vanilla & \sslam & 34 & 50 \\
         \kradar & \sslam & 37 & 60 \\
         \hline
         \vanilla & \rslam & 31 & 60 \\
         \kradar & \rslam & 37 & 60 \\
    \end{tabular}
    \caption{Hyperparameter summary as results of the hyperparameter grid-search optimisation.\label{tab:hyperpar}}
\end{table}

\cref{fig:pr_curves} shows a family of \gls{pr} curves for various methods that it is possible to perform SeqSLAM with radar data.
Here, only a single trajectory pair is considered (one as the map trajectory, the other as the live trajectory).
From~\cref{fig:pr_curves} it is evident that:
\begin{enumerate}
    \item\label{obs:embed_v_scans} Performance when using the learned metric embeddings is superior to either polar or cartesian radar scans,
    \item\label{obs:seq_vs_nn} Sequence-based matching of trajectories outperforms \gls{nn}-based searches,
    \item\label{obs:vanilla_vs_kradar} Performance when using the baseline architecture is outstripped by the rotationally-invariant modifications, and
    \item\label{obs:fwd_vs_bwd} Performance when using the modified search algorithm is boosted.
\end{enumerate}

Observation~\ref{obs:embed_v_scans} can be attributed to the fact that the learned representation is designed to encode only knowledge concerning place, whereas the imagery is subject to sensor artefacts.
Observation~\ref{obs:seq_vs_nn} can be attributed to perceptual aliasing along straight, canyon-like sections of an urban trajectory being mitigated.
Observation~\ref{obs:vanilla_vs_kradar} can be attributed to the rotationally-invariant architecture itself.
Observation~\ref{obs:fwd_vs_bwd} can be attributed to the ability of the adjusted search to detect loop closures in reverse.

\cref{tab:metrics} provides further evidence for these findings by aggregating \gls{pr}-related statistics over the entirety of the dataset discussed in~\cref{sec:experimental:demarc} -- the map trajectory is kept constant and the live trajectory varies over forays spanning a month of urban driving.

While it is clear that we outperform \gls{nn} techniques presented in~\cite{KidnappedRadarArXiv}, the F-scores in~\cref{tab:metrics} present a mixed result when comparing the standard SeqSLAM search and the modified search discussed in~\cref{sec:method:rseqslam}.
However, consider~\cref{fig:score_thr}.
Here, the structure of the backwards loop closures is discovered more readily by the backwards search.

It is important to remember when inspecting the results shown in~\cref{tab:metrics,fig:pr_curves} that the data in this part of the route (c.f.~\cref{sec:experimental:demarc}) is \emph{unseen} by the network during training, and particularly challenging.
This is a necessary analysis of the generalisation of learned place recognition methods but is not a requirement when deploying the learned knowledge in \gls{tr} modes of autonomy.

The takeaway message is that we have improved the recall at good precision levels by about \SI{30}{\percent} by applying sequence-based place recognition techniques to our learned metric space.

\begin{table*}[]
\renewcommand{\arraystretch}{2}
    \centering
    \begin{tabular}{cc|cccccc}
         \textbf{Representation} & \textbf{Search} & \gls{auc} & max $F_{1}$ & max $F_{2}$ & max $F_{0.5}$ & $R_{P = 60\%}$ & $R_{P = 80\%}$\\
         \hline
         \vanilla & \gls{nn} & 0.26 & 0.34 & 0.37 & 0.29 & 0.03 & 0.00\\
         \kradar & \gls{nn} & 0.37 & 0.41 & 0.40 & 0.41 & 0.16 & 0.06\\
         \hline
         \vanilla & \sslam & 0.47 & 0.49 & 0.37 & 0.60 & 0.40 & 0.31\\
         \kradar & \sslam & 0.52 & \textbf{0.53} & 0.41 & \textbf{0.64} & 0.45 & \textbf{0.37}\\
         \hline
         \vanilla & \rslam & 0.46 & 0.48 & 0.36 & 0.60 & 0.39 & 0.32\\
         \kradar & \rslam & \textbf{0.53} & \textbf{0.53} & \textbf{0.42} & 0.62 & \textbf{0.46} & 0.36\\
    \end{tabular}
    \caption{Summary statistics for various radar-only SeqSLAM techniques (including representation of the radar frame and style of search) as aggregated over a month of urban driving.
    All quantities are expressed as a mean value.
    As discussed in~\cref{sec:experimental:metrics}, the requirement imposed on matches (as true/false positives/negatives) is more strict than that presented in~\cite{KidnappedRadarArXiv} with consequentially worse performance than previously published for \vanilla~\kradar, and \nn~systems.
    The key message of this paper is that sequence-based exploitation of these learned metric embeddings (middle and bottom rows) is beneficial in comparison to \gls{nn} matching in a discretised search space (top two rows).}
    \label{tab:metrics}
\end{table*}

\section{Conclusion}
\label{sec:concl}

We have presented an application of recent advances in learning representations for imagery obtained by radar scan formation to sequence-based place recognition.
The proposed system is based on a manipulation of off-the-shelf SeqSLAM with prudent adjustments made taking into account the complete sweep made by scanning radar sensors.
We have have further proven the utility of our rotationally invariant architecture -- a crucial enabling factor of our SeqSLAM variant.
Crucially, we achieve a boost of \SI{30}{\percent} in recall at high levels of precision over our previously published \acrlong{nn} approach.

\section{Future Work}
\label{sec:fut}

In the future we plan to retrain and test the system on the all-weather platform described in~\cite{kyberd2019}, a signficant factor in the development of which was to explore applications of \gls{fmcw} radar to mobile autonomy in challenging, unstructured environments.
We also plan to integrate the system presented in this paper with our mapping and localisation pipeline which is built atop of the scan-matching algorithm of~\cite{2018ICRA_cen,2019ICRA_cen}.

\section*{Acknowledgment}

This project is supported by the Assuring Autonomy International Programme, a partnership between Lloyd’s Register Foundation and the University of York as well as UK EPSRC programme grant EP/M019918/1.
We would also like to thank our partners at Navtech radar.

\bibliographystyle{IEEEtran}
\bibliography{biblio}

\end{document}